%% file: main.tex
\documentclass[10pt,twocolumn,letterpaper]{article}

\usepackage[pagenumbers]{cvpr} 
\usepackage{makecell}
\usepackage[table]{xcolor}
\usepackage[ruled,vlined]{algorithm2e}
\usepackage{booktabs}
\usepackage{multirow}
\usepackage{tabularx}
\usepackage{float}
\usepackage[percent]{overpic}
\usepackage{graphicx}
\usepackage[ruled,vlined]{algorithm2e}

\definecolor{deepgreen}{RGB}{0,100,0}
\input{preamble}
\definecolor{cvprblue}{rgb}{0.21,0.49,0.74}
\usepackage[pagebackref,breaklinks,colorlinks,allcolors=cvprblue]{hyperref}

\def\paperID{17655} 
\def\confName{CVPR}
\def\confYear{2026}

\title{RedVTP: Training-Free Acceleration of 
Diffusion Vision-Language Models
Inference 
via Masked Token-Guided Visual Token Pruning}

\author{
Jingqi Xu\textsuperscript{1}\thanks{Equal contribution.} \quad
Jingxi Lu\textsuperscript{1}\footnotemark[1] \quad
Chenghao Li\textsuperscript{1}\footnotemark[1] \\
Sreetama Sarkar\textsuperscript{1} \quad
Souvik Kundu\textsuperscript{2} \quad
Peter A. Beerel\textsuperscript{1} \\
\textsuperscript{1}University of Southern California \quad
\textsuperscript{2}Intel Labs \\
{\tt\small \{jingqixu, jingxil, cli78217, sreetama, pabeerel\}@usc.edu} \\
{\tt\small mail2ksouvik@gmail.com}
}

\begin{document}
\maketitle
\input{sec/0_abstract}    
\input{sec/1_intro}

\input{sec/2_formatting}
\input{sec/3_finalcopy}
{
    \small
    \bibliographystyle{ieeenat_fullname}
    \bibliography{main}
}

\input{sec/X_suppl}

\end{document}

%% file: sec/0_abstract.tex
\begin{abstract}
Vision-Language Models (VLMs) have achieved remarkable progress in multimodal reasoning and generation, yet their high computational demands remain a major challenge. Diffusion Vision-Language Models (DVLMs) are 
particularly attractive because they enable parallel token decoding, but the large number of visual tokens still significantly hinders their inference efficiency. While visual token pruning has been extensively studied for autoregressive VLMs (AVLMs), it remains largely unexplored for DVLMs. In this work, we propose RedVTP, a response-driven visual token pruning strategy that leverages the inference dynamics of DVLMs. Our method estimates visual token importance using attention from the masked response tokens. Based on the observation that these importance scores remain consistent across steps, RedVTP prunes the less important visual tokens from the masked tokens after the first inference step, thereby maximizing inference efficiency. Experiments show that RedVTP improves token generation throughput of LLaDA-V and LaViDa by up to 186\% and 28.05\%, respectively, and reduces inference latency by up to 64.97\% and 21.87\%, without compromising—and in some cases improving—accuracy. Our code is available at \url{https://github.com/Blacktower27/RedVTP}.
\end{abstract}

%% file: sec/1_intro.tex
\section{Introduction}
\label{sec:intro}
Autoregressive 
vision-language models (AVLMs), such as GPT-5~\cite{openai2025gpt5}, LLaVA~\cite{liu2024improvedbaselinesvisualinstruction,liu2024llavanext}, Qwen-VL~\cite{bai2023qwenvlversatilevisionlanguagemodel}, and Gemini~\cite{comanici2025gemini}, have demonstrated impressive capabilities across a wide range of vision-language tasks~\cite{du2022survey}. These models generate high-quality response text conditioned on image-prompt pairs. 
However, the inherent causal attention mechanism limits the model’s ability to capture bidirectional dependencies, which results in suboptimal performance on tasks requiring reverse reasoning or masked text recovery~\cite{nie2025largelanguagediffusionmodels,li2025lavidalargediffusionlanguage}. For instance, in reverse reasoning, given the conclusion “The sidewalk is wet,” the model is expected to infer a plausible cause like “It rained last night.” In masked text recovery, it must accurately predict missing spans, as in “The [MASK] chased the cat.”
However, the model is restricted during training such that earlier tokens cannot attend to future tokens, leading to degraded performance on these tasks.

\begin{figure}[t]
    \centering
    \includegraphics[width=1\linewidth]{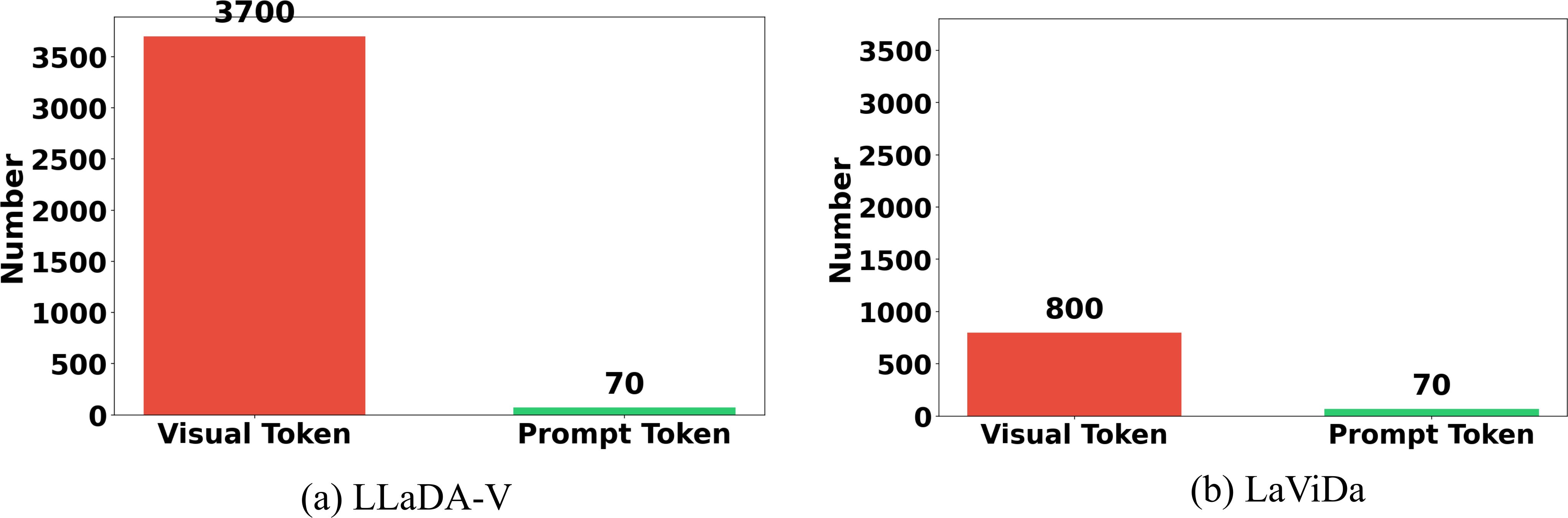}
    \caption{Comparison of the average numbers of visual and prompt tokens on InfoVQA~\cite{mathew2021infographicvqa} using (a) LLaDA-V and (b) LaViDa.}
    \label{fig:comparison}
\end{figure}

Diffusion vision-language models (DVLMs), such as LLaDA-V~\cite{you2025lladavlargelanguagediffusion} and LaViDa~\cite{li2025lavidalargediffusionlanguage}, have recently been proposed. Compared with AVLMs, the key difference lies in the decoding paradigm: instead of generating one response token sequentially at each step, DVLMs decode several response tokens at multiple positions simultaneously in each inference step through an unmasking process that leverages bidirectional attention~\cite{ye2025dream7bdiffusionlarge}. 
While this bidirectional attention enhances DVLMs’ ability to capture bidirectional dependencies, it also prevents the use of traditional key-value (KV) caching~\cite{li2025lavidalargediffusionlanguage} typically used to 
accelerate inference. 
Figure~\ref{fig:comparison} shows that, as in AVLMs, the number of visual tokens in DVLMs is much larger than that of prompt tokens. For example, the average number of visual tokens is approximately 50 times and 10 times greater than the number of prompt tokens per sample on InfoVQA~\cite{mathew2021infographicvqa} when using LLaDA-V and LaViDa, respectively.

Although visual token pruning for DVLMs has not yet been explored,  
it has proven to be effective in accelerating inference for AVLMs.
Training-based methods~\cite{cao2023pumer}~\cite{Li2023BLIP2BL}~\cite{bai2023qwenvlversatilevisionlanguagemodel} typically involve training additional layers to aggregate the visual information from a large number of visual tokens into a small set of learnable queries. However, such approaches incur additional computational overhead for training. For training-free methods~\cite{jiang2024foprufocalpruningefficient}~\cite{arif2024hiredattentionguidedtokendropping}~\cite{yang2024visionzip}~\cite{shang2024llavaprumergeadaptivetokenreduction}, before projecting the large amount of visual tokens into the textual space, they typically use the attention maps extracted from the vision encoder to measure the importance of visual tokens, and either prune the less important ones or aggregate them into a few visual representations.

To fill the gap in visual token pruning methods for DVLMs, 
we propose RedVTP, a training-free, response-driven visual token pruning strategy to accelerate DVLMs inference. 
We leverage the still masked response tokens to measure the importance of visual tokens and
observe that masked token-guided importance scores computed after each inference step exhibit high similarity.
Based on this finding, we prune the less important visual tokens once after the first inference step to maximally 
increase inference efficiency while maintaining generation accuracy. 

To evaluate the effectiveness of our method, we use two pioneering DVLMs, LLaDA-V~\cite{you2025lladavlargelanguagediffusion} and LaViDa~\cite{li2025lavidalargediffusionlanguage}, and conduct experiments on six widely used benchmarks~\cite{fu2025mmecomprehensiveevaluationbenchmark}~\cite{kembhavi2016diagramworthdozenimages}~\cite{mathew2021docvqadatasetvqadocument}~\cite{mathew2021infographicvqa}~\cite{xai2024grok1.5v}~\cite{liu2024mmbenchmultimodalmodelallaround}. Experimental results demonstrate that our RedVTP can significantly accelerate DVLM inference without sacrificing accuracy, and even improve accuracy on certain benchmarks.

Our main contributions are summarized as follows: 1) We propose 
RedVTP, to the best of our knowledge, the first approach that accelerates inference in DVLMs through visual token pruning. 
2) We introduce a response-driven visual token pruning strategy that innovatively leverages still masked tokens to measure importance of visual tokens and prunes less important ones. 3) Extensive experiments on pioneering DVLMs and widely used benchmarks demonstrate that, compared to the original models, RedVTP can improve token generation throughput by up to 186\%, reduce inference latency by up to 64.97\%, without sacrificing much accuracy. Compared with the representative visual token pruning method designed for AVLMs, our method achieves up to 9.15\% improvement in accuracy, demonstrating the superiority of our approach.
    
    

\section{Related Work}
\label{sec:intro}

\subsection{AVLMs}
Closed-source AVLMs, such as GPT-5~\cite{openai2025gpt5}, Gemini 2.5 Pro~\cite{comanici2025gemini}, and Claude 3.7 Vision~\cite{anthropic2025claude37}, have demonstrated impressive performance across a wide range of vision tasks. In parallel, open-source AVLMs have also rapidly advanced, including LLaVA-v1.5~\cite{liu2024improvedbaselinesvisualinstruction}, LLaVA-Next~\cite{liu2024llavanext}, Llama 4~\cite{meta2025llama4}, Maya~\cite{alam2025mayabuildingmultilingualvision}, DeepSeekVL~\cite{lu2024deepseekvlrealworldvisionlanguageunderstanding}, and Qwen2.5-VL~\cite{bai2025qwen25vltechnicalreport}. However, the large number of visual tokens significantly hinders the inference efficiency of AVLMs~\cite{xu2025hivtptrainingfreemethodimprove}. 

Several visual token pruning methods have been proposed to accelerate AVLMs inference. 
In the category of training-based methods, PuMer~\cite{cao2023pumer} inserts token reducer modules before certain transformer layers in the LLM to learn which visual tokens should be merged into fewer ones. 
Q-Former~\cite{Li2023BLIP2BL} and Resampler~\cite{bai2023qwenvlversatilevisionlanguagemodel} train additional cross-attention layers to aggregate information from a large number of visual tokens into a few learnable queries. 
However, training these additional components results in substantial computational overhead.
In the category of training-free methods, 
FoPru~\cite{jiang2024foprufocalpruningefficient} and HiRED~\cite{arif2024hiredattentionguidedtokendropping} utilize attention maps from the vision encoder to compute the average attention score for each visual token as its importance score, and prune the less important ones.
Building upon similar importance score computing strategies, VisionZip~\cite{yang2024visionzip} and PruMerge~\cite{shang2024llavaprumergeadaptivetokenreduction} perform average merging among less important visual tokens and preserve the resulting merged representations.

\subsection{DVLMs}
The causal attention mechanism limits the ability of AVLMs to perform reverse reasoning or recover masked text. To overcome these limitations, DVLMs have recently developed rapidly, such as LLaDA-V~\cite{you2025lladavlargelanguagediffusion} and LaViDa~\cite{li2025lavidalargediffusionlanguage}. In the LLM component, they adopt a bidirectional attention mechanism in the transformer layers, which allows all tokens to interact with each other. Conditioned on visual and text tokens, a certain number of response tokens are decoded from the masked positions at each inference step. However, the large number of visual tokens, coupled with the inability to leverage traditional KV caching due to the bidirectional attention mechanism, results in significantly low inference efficiency for DVLMs. To the best of our knowledge, the problem of accelerating DVLMs inference through visual token pruning remains underexplored.

%% file: sec/2_formatting.tex
\begin{figure*}[t]
    \centering
    \includegraphics[width=0.9\linewidth]{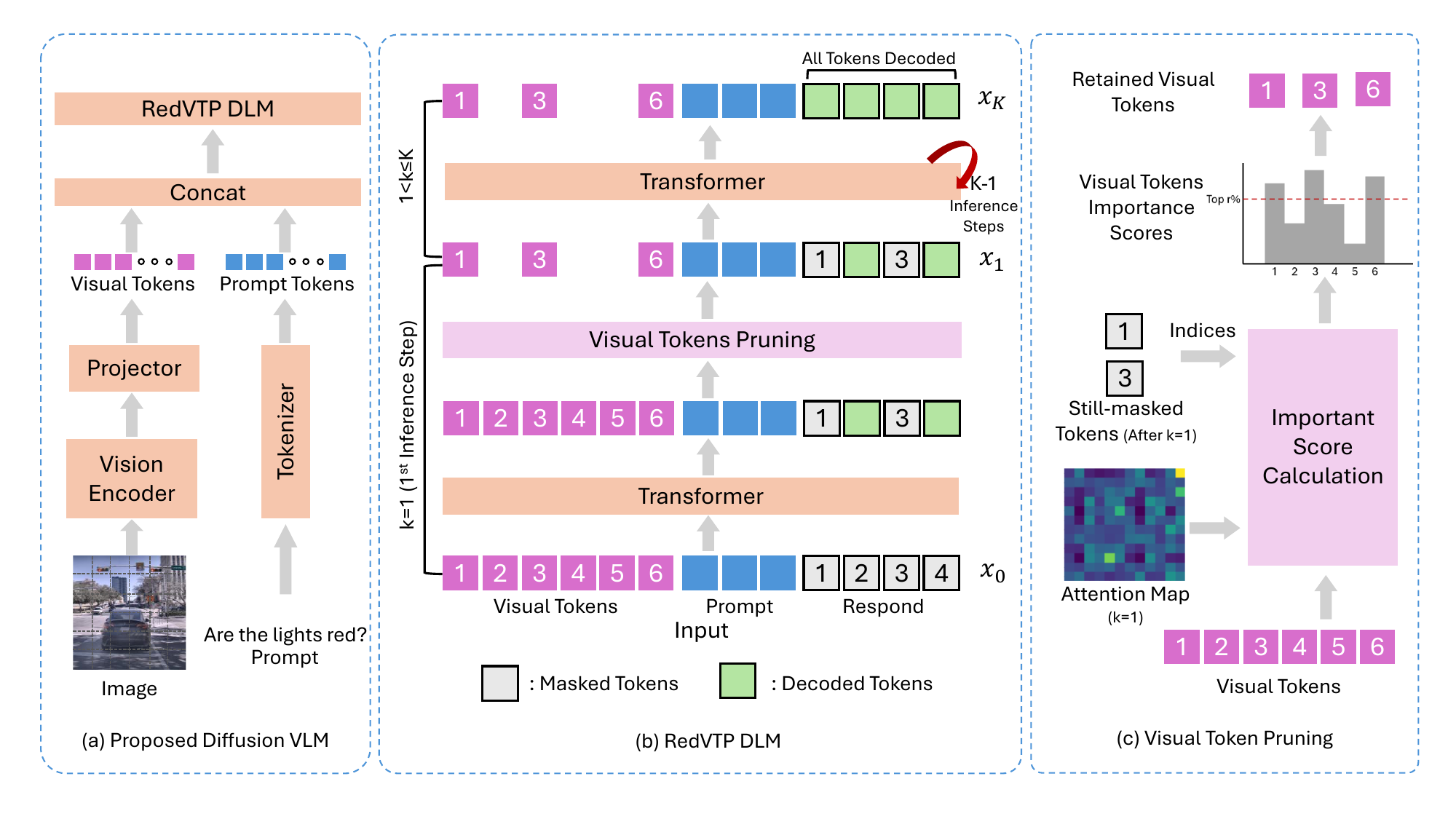}
    \caption{The framework of RedVTP on DVLMs. a) The proposed DVLM process: We apply RedVTP on the original diffusion language model. b) DVLM with RedVTP applied after the $1^{st}$ inference step. c) We collect the attention map average on heads and layers from the $1^{st}$ inference step and calculate the cumulative attention received from all still-masked response tokens of each image tokens to get their importance scores. Based on the importance scores, we select top-\( r \) proportion of visual tokens to retain for the remaining inference steps.}
    \label{fig:framework}
\end{figure*}

\section{Preliminaries}
\label{sec:Pre}
\noindent\textbf{Architecture of DVLMs.}
DVLMs generate responses of fixed length 
$\tau$
based on an image-text pair. Typically, a DVLM consists of three components: a vision encoder, a projector, and a diffusion language model (DLM)~\cite{nie2025largelanguagediffusionmodels,ye2025dream7bdiffusionlarge}. The vision encoder divides the input image into $N$ patches, each of which is linearly mapped to a $d_v$-dimensional embedding. These $N$ visual tokens are then processed by a set of transformer layers in the vision encoder to enable visual information interaction. Subsequently, a projector maps the resulting embeddings into the language space to align the visual and textual modalities 
, producing $V \in \mathbb{R}^{N \times d}$, where $d$ denotes the shared embedding dimension across modalities. Simultaneously, the input prompt is tokenized and embedded into textual embeddings $T \in \mathbb{R}^{m \times d}$, where $m$ represents prompt length. The DLM $P_\theta(\cdot)$ consists of $L$ transformer layers, each with $H$ attention heads, which iteratively decode the masked response tokens based on the joint context of $V$ and $T$ through $K$ inference steps. 
Specifically, we denote the initial response token sequence, where all tokens are masked, as \( R_1 = [M, \dots, M] \in \mathbb{R}^{\tau \times d} \), where \( M \in \mathbb{R}^d \) represents the embedding of a 
masked token. At the \( k \)-th inference step, where \( 1 \leq k \leq K \), the input to the DLM is denoted as $X_k = Concat(V, T, R_k)$, and the inference step is formulated as:
\begin{equation}
q(X_{k+1} \mid X_k) = \prod_{i=1}^{N+m+\tau} q(X^i_{k+1} \mid X_k)
\label{eq:transition_step}
\end{equation}
where \( q(X^i_{k+1} \mid X_k) \) denotes the probability distribution over the possible values that the \( i \)-th token may take, and \( q(X_{k+1} \mid X_k) \) represents the joint probability distribution over all tokens. Specifically, if \( X^i_k \neq M \) (
meaning tokens from \( V \), \( T \), and already decoded response tokens), then the value of \( X^i_{k+1} \) remains unchanged with probability 1:
\begin{equation}
\footnotesize
q(X^i_{k+1} \mid X_k) = 1, \quad X^i_k \neq M,\; X^i_{k+1} = X^i_k
\label{eq:token_copy}
\end{equation}
Otherwise, if \( X^i_k = M \), the probability that \( X^i_{k+1} \) remains still masked is denoted by \( q_k \), and the probability of being decoded into a meaningful token predicted by \( P_\theta(\cdot) \) is \( (1 - q_k) P_\theta(X^i_K \mid X_k) \). This can be formulated as:
\begin{equation}
\footnotesize
q(X^i_{k+1} \mid X_k) = 
\begin{cases}
q_k = \frac{1 - \frac{k}{K}}{1 - \frac{k-1}{K}}, & X^i_k = M,\; X^i_{k+1} = M, \\
(1 - q_k) P_\theta(X^i_K \mid X_k), & X^i_k = M,\; X^i_{k+1} \neq M,
\end{cases}
\label{eq:token_transition}
\end{equation}
Then, \( X_{k+1} \) serves as the input to the \( (k+1) \)-th inference step.

\noindent\textbf{Computation Complexity.}
 For each transformer layer in \( P_\theta(\cdot) \), it consists of two key components, including the multi-head self-attention mechanism and a feed-forward network (FFN). We denote the intermediate size of the FFN as \( \mu \). The total floating-point operations (FLOPs) of DVLMs can be expressed as:
\begin{equation}
\text{FLOPs} = L \times K \times \left( 4nd^2 + 2n^2d + 2nd\mu \right),
\end{equation}
where \( n = N + m + \tau \) denotes the input sequence length at each inference step. This equation reveals that the computational complexity is strongly influenced by the sequence length \( n \). Since the number of visual tokens \( N \) significantly exceeds that of textual and response tokens, pruning less important visual tokens is an effective way to accelerate the inference process.

\section{Method}
\subsection{Overview}
The framework of RedVTP is illustrated in Figure~\ref{fig:framework}. 
After the first inference step, we use the masked token-guided importance score to measure the importance of each visual token.
Then, we only retain visual tokens with high importance scores to improve the model's inference efficiency. The algorithmic details are presented in Appendix~\ref{subsec:alg}.


\subsection{Masked Token-Guided Importance Score}
\noindent\textbf{Motivation.}
The inference mechanism of DVLMs allows us to access the attention from still masked response tokens to visual tokens after each inference step.
The decoded response tokens have already been determined, and the visual information they attend to is no longer relevant because their prediction values cannot be changed. In contrast, the still masked response tokens have not yet been determined, and their prediction values can further affect the accuracy of the final response. Therefore, only the visual tokens attended by the still masked response tokens can influence the final response.
We therefore hypothesize that for DVLMs, the importance scores of visual tokens should be measured 
based on the attention received from the still masked tokens. 
We provide additional support for this hypothesis in Section~\ref{sec:abl} in which we explore using other sets of tokens to guide pruning but get inferior results.
\begin{figure}[t]
    \centering
    \includegraphics[width=1\linewidth]{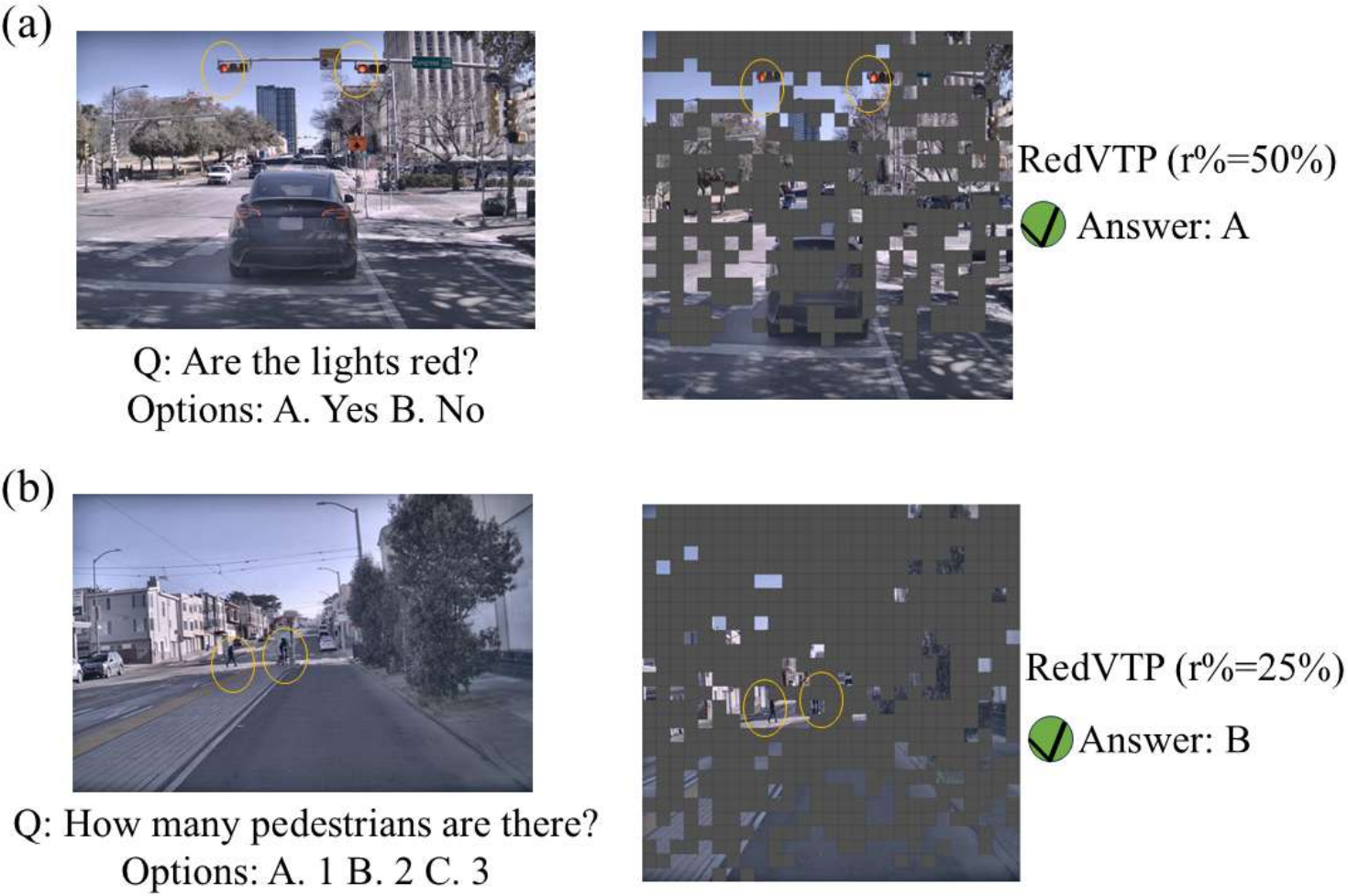}
    \caption{Visualization of visual token pruning results based on masked token-guided importance scores using examples from RealworldQA. 
    The yellow circled regions indicate the areas that the correct responses should attend to. 
    (a) under \( r = 50\% \); (b) under \( r = 25\% \). 
    As can be observed, the regions essential for generating the correct responses are consistently retained under different settings.}
    \label{fig:realworld}
\end{figure}

\noindent\textbf{Method description.} At the \( k \)-th inference step, we denote the attention map of the \( l \)-th transformer layer and the \( h \)-th head in the \( P_\theta(\cdot) \) as \(\mathbf{A}_k^{\mathrm{(l,h)}} \in \mathbb{R}^{(N + m + \tau) \times (N + m + \tau)}\), where \( 1 \leq l \leq L \) and \( 1 \leq h \leq H \). In the output of the \( k \)-th inference step, \( X_{k+1} \), 
we use \( \mathcal{I}(V) \) to denote the indices of visual tokens, \( \mathcal{I}_k(\text{M}) \) to denote the indices of still masked response tokens, and \( \mathcal{I}_k(\text{D}) \) to denote the indices of decoded response tokens.
The visual token importance score after the \( k \)-th inference step is formulated as:
\begin{equation}
\small
\bar{\mathbf{A}} = \frac{1}{H} \frac{1}{L} \sum_{h=1}^{H} \sum_{l=1}^{L} \mathbf{A}_k^{(l,h)},
\label{eq:attn_avg}
\end{equation}
\begin{equation}
\mathbf{S_k} = \frac{1}{|\mathcal{I}_k(\text{M})|} \sum_{j \in \mathcal{I}_k(\text{M})} \bar{\mathbf{A}}_{j, \mathcal{I}(V)},
\label{eq:importance_score}
\end{equation}
where \( \mathbf{S}_k \in \mathbb{R}^{N} \) denotes the averaged importance scores of \( N \) visual tokens with respect to the still masked response tokens. 
Using LLaDA-V~\cite{you2025lladavlargelanguagediffusion} on samples from InfoVQA~\cite{mathew2021infographicvqa}, 
we observe that the importance scores across all inference steps exhibit a high degree of similarity, indicating that the visual tokens deemed important with respect to the still masked response tokens remain largely consistent, as shown in Figure~\ref{fig:importance similarity}.
The detailed experimental description is provided in Section~\ref{sec:exp}.

\subsection{Visual Pruning}
Based on our finding that the visual token importance scores remain nearly unchanged across all inference steps, and to run full-sequence inference as few times as possible, we retain a proportion \( r \) of visual tokens with the highest importance scores according to \( S_1 \) once after the first inference step.
In this way, we only need to do the heavy full sequence inference once.
The process can be formulated as:
\begin{equation}
\mathcal{I}^{\text{keep}} = \texttt{Top}(\mathcal{I}(V), \mathbf{S}_1, r),
\end{equation}
where \( \texttt{Top}(\cdot) \) selects the subset of \( \mathcal{I}(V) \) corresponding to the top-\( r \) proportion of visual tokens with the highest importance scores.
We denote the total number of retained important visual tokens as \( N_r = |\mathcal{I}^{\text{keep}}| \),  
where \( |\mathcal{I}^{\text{keep}}| = N \cdot r \). 
Based on \( \mathcal{I}^{\text{keep}} \), we extract the corresponding visual tokens from \( V \in \mathbb{R}^{N \times d} \) and arrange them in their original order to form the retained set \( V^{\text{keep}} \in \mathbb{R}^{N_r \times d} \).
After this visual token pruning process, \( V^{\text{keep}} \) will be used as the visual input of \( X_k \)  
for \( 1 < k \leq K \), completing the subsequent inference process. 

\section{Experimental Results}
\label{sec:exp}
In this section, we first present a detailed analysis comparing \( S_1 \) and \( S_k \) (\(1 < k \leq K{-}1\)) to evaluate their similarity.
In addition, we systematically evaluate RedVTP on six widely used benchmarks using two state-of-the-art DVLMs: LLaDA-V~\cite{you2025lladavlargelanguagediffusion} and LaViDa~\cite{li2025lavidalargediffusionlanguage}. 
In order to improve inference efficiency, LaViDa has already applied a \(2 \times 2\) average pooling operation to the visual tokens after the vision encoder, retaining only 25\% of the original tokens. 
We compare LaViDa with our proposed method applied to LLaDA-V under the same 25\% token retention ratio.
Notably, LLaDA-V and LaViDa share the same architectural components (for example, both use SigLIP~\cite{zhai2023sigmoidlosslanguageimage} as the vision encoder and LLaDA~\cite{nie2025largelanguagediffusionmodels} as the LLM).
\subsection{Experimental Setup}
We adopt six widely used benchmarks, including: 
1) Ai2D~\cite{kembhavi2016diagramworthdozenimages}, for evaluating the ability to understand and answer questions about scientific diagrams; 
2) DocVQA~\cite{mathew2021docvqadatasetvqadocument}, for evaluating document understanding and visual question answering from scanned documents; 
3) RealworldQA~\cite{xai2024grok1.5v}, for assessing model performance on real-world images with natural scene complexities; 
4) InfoVQA~\cite{mathew2021infographicvqa}, for evaluating comprehension and reasoning over infographic-style images; 
5) MME~\cite{fu2025mmecomprehensiveevaluationbenchmark}, for evaluating visual perception and recognition abilities; and 
6) MMBench(MMB)~\cite{liu2024mmbenchmultimodalmodelallaround}, for evaluating multimodal understanding and reasoning capabilities.
When evaluating RedVTP on LLaDA-V, we conduct experiments under \( r \in \{75\%, 50\%, 25\%\} \). 
We also demonstrate results for further token compression using our approach on top of the 75\% token reduction performed by LaViDa.
When applying RedVTP with \( r = 75\% \), only 18.75\% of the original visual tokens remain. 
For LaViDa, we use the same hyperparameters as in LLaDA-V across all benchmarks (e.g., generation length, total inference steps). All experiments are conducted using an NVIDIA RTX A6000 GPU.
\subsection{Step-wise Comparison of Importance Scores}
\begin{figure}[t]
    \centering
    \includegraphics[width=1\linewidth]{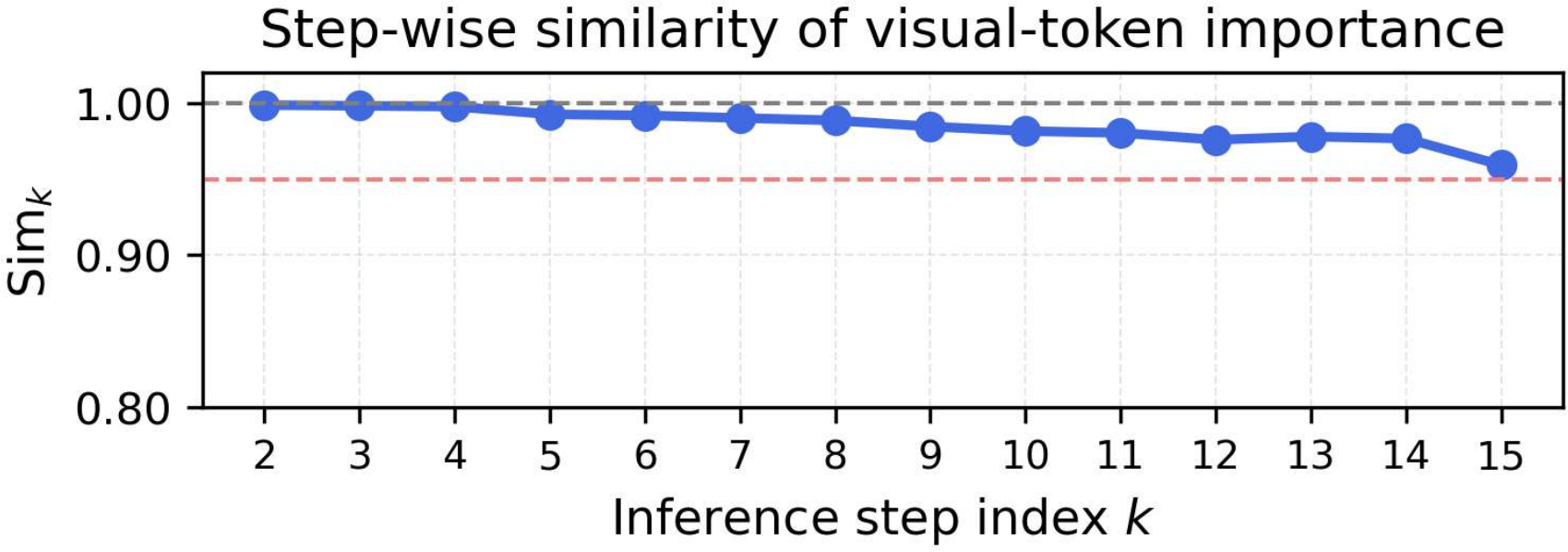}
    \caption{The averaged masked token-guided importance score similarity between \( S_1 \) and each \( S_k \) (\(1 < k \leq 15\)), 
    computed using 20\% of the samples from InfoVQA with LLaDA-V. 
    \( \text{Sim}_k \) denotes the cosine similarity between \( S_k \) and \( S_1 \). 
    From the curve, we can observe that all \( \text{Sim}_k \) values are higher than 0.95.}
    \label{fig:importance similarity}
\end{figure}

To compare the visual token importance scores \( S_1 \), obtained after the first inference step, and \( S_k \) (\(1 < k \leq K{-}1\)), obtained from subsequent inference steps, we randomly sample 20\% of instances from InfoVQA and conduct the analysis on LLaDA-V. The total number of inference steps is set to \( K = 16 \), following the default configuration of LLaDA-V on InfoVQA. For each instance, we compute the cosine similarity between \( S_1 \) and each \( S_k \) (\(1 < k \leq 15\)), and report the average similarity across all samples, denoted as \( \text{Sim}_k \) (\(1 < k \leq 15\)). Since after the 16-th inference step all tokens have been fully decoded, the maximum value of \( k \) considered here is 15.
The averaged similarity results for each \( \text{Sim}_k \) are visualized in Figure~\ref{fig:importance similarity}. As we can observe, all \( \text{Sim}_k \) values are greater than 0.95. 
These results demonstrate that the attention from the still-masked response tokens to the visual tokens remains highly consistent across all inference steps.
\subsection{Accuracy Analysis}
In Table~\ref{tab:accuracy_combined}, we present the accuracy of our proposed method applied to LLaDA-V and LaViDa across six benchmarks.

\noindent\textbf{Accuracy on LLaDA-V.}
As shown in Table~\ref{tab:accuracy_combined}, compared with the original LLaDA-V, when the visual token retaining ratio is \( r= 75\% \), our method even achieves an average accuracy that is 0.16\% higher across all benchmarks. This is because our method outperforms the original LLaDA-V on Ai2D, DocVQA, and MME by 1.28\%, 0.39\%, and 1.22\%, respectively. When the retaining ratio is \( r = 50\% \), our method results in an average accuracy drop of only 0.26\% compared with the original LLaDA-V across all benchmarks. It is worth noting that on Ai2D and RealworldQA, our method still achieves higher accuracy than the original LLaDA-V by 0.68\% and 0.93\%, respectively. Even at a low retaining ratio of \( r = 25\% \), our method still preserves accuracy well, with the average accuracy across all benchmarks being only 4.15\% lower than that of the original LLaDA-V.

\noindent\textbf{Accuracy on LaViDa.} As shown in Table~\ref{tab:accuracy_combined}, when the visual token retaining ratio is \( r= 75\% \), compared with the original LaViDa, our proposed method achieves an average accuracy that is only 2.2\% lower across these benchmarks. 
This demonstrates that our RedVTP can achieve further token pruning in LaViDa.
We hypothesize that the possible reason is that although visual token compression spatially reduces the number of visual tokens, it does not analyze the importance of visual tokens from a semantic perspective. As a result, within the compressed visual tokens, there may still exist some tokens that are less important to the still masked response tokens.

\begin{figure*}[t]
    \centering
    \includegraphics[width=0.9\linewidth]{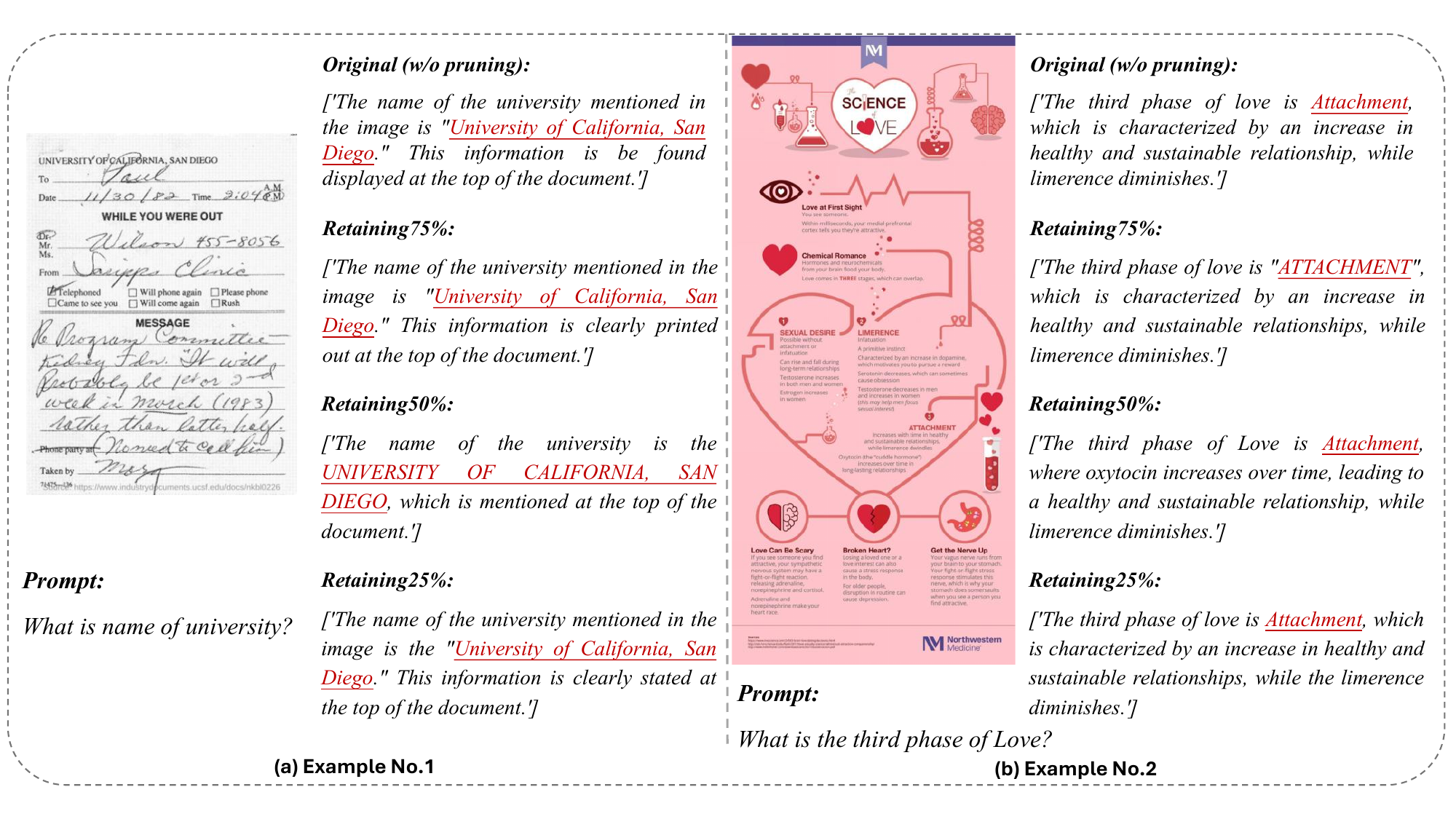}
    \caption{Two examples from DocVQA and InfoVQA using LLaDA-V. It can be observed that the model can correctly answer the questions even with only $25\%$ visual tokens being retained.}
    \label{fig:Visual_examples}
\end{figure*}

\noindent\textbf{LaViDa vs. LLaDA-V+RedVTP.}
We compare LaViDa and LLaDA-V+RedVTP (\(r = 25\%\)) in terms of accuracy. Our method achieves an average accuracy that is 23.72\% higher across these benchmarks.
In particular, on DocVQA and InfoVQA, the accuracy of our method exceeds that of the LaViDa by 28.40\% and 78.70\%, respectively, which demonstrates that our method maintains accuracy better than the visual token compression approach when retaining the same number of visual tokens. We hypothesize that a possible reason is that visual token compression in LaViDa reduces the number of visual tokens via local averaging, which may blur local detailed information. This leads to a significant performance drop on benchmarks that require fine-grained visual understanding, such as DocVQA and InfoVQA.

\noindent\textbf{Discussion.}
Overall, on both LLaDA-V and LaViDa, our proposed method can maintain accuracy with fewer visual tokens across all benchmarks. Two examples are shown in Figure~\ref{fig:Visual_examples}, where the model can still correctly answer the questions even when only 25\% of the visual tokens are retained. 

Under \( r = 75\% \) and \( r = 50\% \) on LLaDA-V, our method even achieves higher accuracy than the original LLaDA-V on Ai2D. This is likely because the task relies 
on a small fraction of the visual information in the frame,
and pruning unimportant visual tokens may make the model focus more on the text prompt. As a result, it can, to some extent, actually improve performance.

\begin{table*}[ht]
\centering
\begingroup
\footnotesize
\renewcommand{\arraystretch}{1.3}
\begin{tabular}{l|c|ccc|c|c}
\toprule
\multicolumn{7}{c}{\textbf{Accuracy Performance Comparison}} \\
\midrule
\multirow{2}{*}{\textbf{Model \& Method}} 
& \multirow{2}{*}{\textbf{LLaDA-V}} 
& \multicolumn{3}{c|}{\textbf{+RedVTP (LLaDA-V)}} 
&\multirow{2}{*}{\textbf{LaViDa}}
& \textbf{+RedVTP (LaViDa)} \\
& & $r=75\%$ & $r=50\%$ & $r=25\%$ & & $r=75\%$ \\
\midrule
\# Vision Tokens (avg.) & 3340 & 2505 & 1670 & 835 & 835 & 626 \\
\midrule
AI2D~\cite{kembhavi2016diagramworthdozenimages}        & 76.29 & \textbf{77.27} & 76.81 & 73.58 & 69.59 & 68.65 \\
DocVQA~\cite{mathew2021docvqadatasetvqadocument}      & 83.90 & \textbf{84.23} & 83.70 & 81.63 & 63.55 & 53.12 \\
RealworldQA~\cite{xai2024grok1.5v} & 63.20 & 62.88 & \textbf{63.79} & 62.89 & 53.46 & 51.63 \\
InfoVQA~\cite{mathew2021infographicvqa}     & \textbf{66.09} & 65.55 & 65.10 & 63.21 & 35.37 & 31.05 \\
MME~\cite{fu2025mmecomprehensiveevaluationbenchmark}         & 1506.50 & \textbf{1525.00} & 1502.65 & 1399.55 & 1342.17 & 1325.87 \\
MMB~\cite{liu2024mmbenchmultimodalmodelallaround}         & \textbf{83.33} & 82.82 & 82.30 & 77.66 & 72.08 & 70.62 \\
\bottomrule
\end{tabular}
\endgroup
\caption{Accuracy performance comparison of RedVTP applied to LLaDA-V and LaViDa across six benchmarks. RedVTP is evaluated at multiple visual token retaining ratios $r \in \{75\%, 50\%, 25\%\}$ for LLaDA-V and $r \in \{75\%\}$ for LaViDa. The row \# Vision Tokens reports the average number of visual tokens retained across all benchmarks under each setting.}
\label{tab:accuracy_combined}
\end{table*}

\begin{table*}[ht]
\centering
\label{tab:efficiency_combined}
\resizebox{\textwidth}{!}{%
\setlength{\tabcolsep}{8pt}
\renewcommand{\arraystretch}{1.2}
\begin{tabular}{llcccccc}
\toprule
\multicolumn{8}{c}{\textbf{Efficiency}} \\
\textbf{Model \& Method} &  & AI2D~\cite{kembhavi2016diagramworthdozenimages} & DocVQA~\cite{mathew2021docvqadatasetvqadocument} & RealworldQA~\cite{xai2024grok1.5v} & InfoVQA~\cite{mathew2021infographicvqa} & MME~\cite{fu2025mmecomprehensiveevaluationbenchmark} & MMB~\cite{liu2024mmbenchmultimodalmodelallaround} \\
\midrule
\multirow{2}{*}{LLaDA-V}
 & Latency (s/sample)~$\downarrow$       & 2.285 & 17.388 & 3.721 & 17.308 & 3.176 & 2.294 \\
 & Throughput (tok/s)~$\uparrow$         & 0.438 & 1.783  & 0.269 & 1.791  & 0.315 & 0.395 \\
\cmidrule(l){2-8}
\multirow{2}{*}{\makecell{+RedVTP\\($r=75\%$)}} 
 & Latency (s/sample)~$\downarrow$       & 2.030 & 13.113 & 2.963 & 13.151 & 2.182 & 1.665 \\
 & Throughput (tok/s)~$\uparrow$         & 0.493 & 2.264  & 0.338 & 2.357  & 0.458 & 0.601 \\
\cmidrule(l){2-8}
\multirow{2}{*}{\makecell{+RedVTP\\($r=50\%$)}} 
 & Latency (s/sample)~$\downarrow$       & 1.836 & 9.725  & 2.746 & 9.601  & 1.990 & 1.823 \\
 & Throughput (tok/s)~$\uparrow$         & 0.545 & 3.187  & 0.364 & 3.229  & 0.502 & 0.549 \\
\cmidrule(l){2-8}
\multirow{2}{*}{\makecell{+RedVTP\\($r=25\%$)}} 
 & Latency (s/sample)~$\downarrow$       & \textbf{1.622} & \textbf{6.377}  & \textbf{2.307} & \textbf{6.062}  & \textbf{2.136} & \textbf{1.391} \\
 & Throughput (tok/s)~$\uparrow$         & \textbf{0.617} & \textbf{4.861}  & \textbf{0.433} & \textbf{5.114}  & \textbf{0.468} & \textbf{0.719} \\
\bottomrule
\end{tabular}%
}
\caption{Latency and throughput of RedVTP applied to LLaDA-V across six benchmarks with visual token retaining ratios \( r \in \{75\%, 50\%, 25\%\} \).}
\label{tab:efficiency_lladav}
\end{table*}

\begin{table*}[ht]
\centering
\label{tab:efficiency_combined}
\resizebox{\textwidth}{!}{%
\setlength{\tabcolsep}{8pt}
\renewcommand{\arraystretch}{1.3}
\begin{tabular}{llcccccc}
\toprule
\multicolumn{8}{c}{\textbf{Efficiency}} \\
\textbf{Model \& Method} &  & AI2D~\cite{kembhavi2016diagramworthdozenimages} & DocVQA~\cite{mathew2021docvqadatasetvqadocument} & RealworldQA~\cite{xai2024grok1.5v} & InfoVQA~\cite{mathew2021infographicvqa} & MME~\cite{fu2025mmecomprehensiveevaluationbenchmark} & MMB~\cite{liu2024mmbenchmultimodalmodelallaround} \\
\midrule
\multirow{2}{*}{LaViDa}
 & Latency (s/sample)~$\downarrow$       & 1.115 & 13.922 & 1.209 & 11.607 & 1.158 & 1.000 \\
 & Throughput (tok/s)~$\uparrow$         & 0.897 & 2.228  & 0.827 & 2.671  & 0.863 & 1.000 \\
\cmidrule(l){2-8}
\multirow{2}{*}{\makecell{+RedVTP\\($r=75\%$)}} 
 & Latency (s/sample)~$\downarrow$       & \textbf{1.035} & \textbf{10.877} & \textbf{1.130} & \textbf{9.624} & \textbf{1.089} & \textbf{0.946} \\
 & Throughput (tok/s)~$\uparrow$         & \textbf{0.966} & \textbf{2.853}  & \textbf{0.885} & \textbf{3.224} & \textbf{0.918} & \textbf{1.058} \\
\bottomrule
\end{tabular}%
}
\caption{Latency and throughput of RedVTP applied to LaViDa across six benchmarks with visual token retaining ratios \( r \in \{75\%\} \).}
\label{tab:efficiency_lavida}
\end{table*}

\subsection{Efficiency Analysis}
To further quantify the inference efficiency advantage of our method, we adopt two widely used evaluation metrics: inference latency (latency)~\cite{li2025lavidalargediffusionlanguage} and token generation throughput (throughput)~\cite{arif2024hiredattentionguidedtokendropping}. Latency measures the total time the model takes to generate a complete response, while throughput measures the number of tokens generated by the model per second. 
The results are shown in Table~\ref{tab:efficiency_lladav} and Table~\ref{tab:efficiency_lavida}, respectively.

As shown in Table~\ref{tab:efficiency_lladav}, when our method is applied to LLaDA-V under \( r = 75\% \), it reduces the average latency across these benchmarks by 23.11\% and increases the average throughput by 32.35\% compared to the original LLaDA-V. Notably, the largest latency reduction of 31.29\% and the highest throughput improvement of 45.50\% are observed on MME. When \( r = 50\% \), our method achieves greater efficiency gains, reducing the average latency by 32.04\% and increasing the average throughput by 52.75\% across all benchmarks. Among them, InfoVQA shows the most significant improvement, with a 44.50\% reduction in latency and an 80.26\% increase in throughput. When \( r = 25\% \), our method reduces the average latency across these benchmarks by 44.57\% and increases the average throughput by 91.66\% compared to the original LLaDA-V. Among the benchmarks, the most significant improvement is observed on InfoVQA, where the latency is reduced by 64.97\% and the throughput is increased by 186\%.

Compared to the LaViDa, when \( r = 75\% \), our method reduces the average latency across all benchmarks by 10.70\% and increases the average throughput by 12.61\%. Particularly, the most significant improvement is observed on DocVQA, where the latency is reduced by 21.87\% and the throughput is increased by 28.05\%. In addition, compared to the original LaViDa, when applying our method on LLaDA-V under \( r = 25\% \), the average throughput across these benchmarks is improved by 57.50\%. 

On both LLaDA-V and LaViDa, our method consistently improves inference efficiency across all benchmarks under different visual token retaining ratios. Specifically, for tasks with long generation lengths such as InfoVQA and DocVQA, the efficiency advantage of our method is particularly significant. This is because these tasks require more inference steps, providing more opportunities for the model to perform reasoning with fewer visual tokens.
\section{Ablation}
\label{sec:abl}
In this study, we evaluate the effectiveness of the two key components in our method: the masked token-guided visual token importance scoring strategy and the visual token pruning strategy. We conduct experiments on LLaDA-V using the RealworldQA, DocVQA, and InfoVQA benchmarks. The visual token retaining ratio is set to \( r = 25\% \) when evaluating the masked token-guided importance scoring strategy, and \( r = 50\% \) when evaluating the visual token pruning strategy.

\noindent\textbf{Justifying Masked Token-Guided Importance Score.}
In our proposed method, after the first inference step, we use the still masked response tokens \( \mathcal{I}_1(\text{M}) \) to measure the visual token importance scores, 
as described in Equations~\ref{eq:attn_avg} and~\ref{eq:importance_score}. Instead of using still masked response tokens, we alternatively compute the visual token importance scores using the attention maps from the first inference step with the following token sets: prompt tokens \( \mathcal{I}_1(\text{T}) \), decoded response tokens \( \mathcal{I}_1(\text{D}) \), all response tokens \( \mathcal{I}_1(\text{M+D}) \) (including both still masked and decoded response tokens), prompt and all response tokens \( \mathcal{I}_1(\text{T+M+D}) \), visual tokens \( \mathcal{I}_1(\text{V}) \), and prompt and still masked response tokens \( \mathcal{I}_1(\text{T+M}) \). 
Furthermore, we also compare our method with FoPru~\cite{jiang2024foprufocalpruningefficient}, a representative pruning method for AVLMs, which uses attention maps from the vision encoder to measure visual token importance.

As shown in Table~\ref{tab:ablation}, using still masked response tokens to compute the visual token importance scores achieves better performance than all other methods across all benchmarks, which demonstrates the superiority of our approach. Specifically, among these methods, FoPru and the one using \( \mathcal{I}_1(\text{V}) \) perform relatively poorly, and our method achieves 6.54\% and 4.26\% higher average accuracy than them, respectively. Compared with the methods using \( \mathcal{I}_1(\text{T}) \), \( \mathcal{I}_1(\text{D}) \), \( \mathcal{I}_1(\text{M+D}) \), \( \mathcal{I}_1(\text{T+M+D}) \), and \( \mathcal{I}_1(\text{T+M}) \), our method achieves 2.39\%, 2.60\%, 1.22\%, 2.00\%, and 1.57\% higher average accuracy across these benchmarks, respectively. The superiority of our method for computing visual token importance scores over other strategies may lie in the fact that, after the first inference step, the final response accuracy is only affected by the prediction quality of the still masked response tokens. Pruning visual tokens that are unimportant to the still masked response tokens does not significantly harm accuracy. In contrast, pruning visual tokens deemed unimportant to other tokens may inadvertently remove several visual tokens that are crucial for the still masked response tokens, leading to a drop in accuracy.

We visualize examples from RealworldQA that are pruned using the masked token-guided importance scores in Figure~\ref{fig:realworld}. As we can observe, when \( r = 50\% \) and \( r = 25\% \), the retained important visual tokens correspond well to the regions that the correct response should attend to.

\noindent\textbf{Pruning Strategy.}
We compare our pruning strategy with random pruning (RP) and progressive pruning (PP). Specifically, for RP, after the first inference step, we randomly retain \( r\) of the visual tokens and prune the remaining ones. A key limitation of iterative pruning in DVLMs is that once a visual token is removed at step $k$, it no longer participates in the input at step $k+1$ and thus cannot be re-evaluated. This irreversibility prevents the model from computing global importance scores across all visual tokens at later steps. 
To address this, we design a PP across all inference steps strategy where, instead of pruning 50\% of visual tokens once after the first inference step, the total 50\% of pruned visual tokens are evenly distributed across all inference steps.
At each step, we recompute importance scores over the current visual tokens and prune a fixed number.

As shown in Table~\ref{tab:ablation}, our pruning strategy consistently performs better across all benchmarks. Specifically, our method outperforms RP by 6.08\%, 27.39\%, and 35.96\% on RealWorldQA, DocVQA, and InfoVQA, respectively. 
The PP achieves similar accuracy to our pruning strategy across these benchmarks. However, PP introduces higher latency (up to 34.36\% slower) and lower throughput (up to 25.49\% lower) due to repeated pruning and shrinking context. The detailed efficiency comparison between our pruning strategy and PP is provided in the Appendix~\ref{subsec:efficiencyPP}.

\begin{table}[t]
\centering
\renewcommand{\arraystretch}{1.1}
{\fontsize{7pt}{8.5pt}\selectfont
\setlength{\tabcolsep}{0.9pt}
\begin{tabularx}{\linewidth}{@{}X c c c@{}}
\toprule
\textbf{Choice} & \textbf{RealworldQA}~\cite{xai2024grok1.5v} & 
\textbf{DocVQA}~\cite{mathew2021docvqadatasetvqadocument} &
\textbf{InfoVQA}~\cite{mathew2021infographicvqa} \\
\midrule

\multicolumn{4}{l}{\textbf{Importance scoring strategy}} \\
\multicolumn{4}{l}{$(r=25\%)$} \\

\textbf{Ours} & \textbf{62.89} & \textbf{81.63} & \textbf{63.21} \\
\( \mathcal{I}_1(\text{T}) \) & 61.44 & 81.27 & 60.55 \\
\( \mathcal{I}_1(\text{D}) \) & 62.27 & 80.66 & 59.82 \\
\( \mathcal{I}_1(\text{M+D}) \) & 61.96 & 81.52 & 61.99 \\
\( \mathcal{I}_1(\text{T+M+D}) \) & 61.83 & 81.57 & 60.69 \\
\( \mathcal{I}_1(\text{V}) \) & 60.04 & 80.10 & 59.23 \\
\( \mathcal{I}_1(\text{T+M}) \) & 62.22 & 80.72 & 61.64 \\
FoPru & 59.72 & 77.61 & 57.91 \\
\midrule

\textbf{Pruning Strategy} & & & \\
$(r=50\%)$ & & & \\

\textbf{Ours} & \textbf{63.79} & \textbf{83.70} & 65.10 \\
RP & 60.13 & 65.70 & 47.88 \\
PP & 63.79 & 83.17 & \textbf{65.96} \\
\bottomrule
\end{tabularx}
}
\caption{Ablation study on importance scoring and pruning strategies on LLaDA-V.}
\label{tab:ablation}
\end{table}



\section{Conclusions and Future Work}
In this paper, we introduce a training-free method, RedVTP, to the best of our knowledge, the first approach to accelerate the inference of DVLMs through visual token pruning. Benefiting from the unique inference mechanism of DVLMs, we utilize masked tokens to measure the importance of visual tokens and prune the less important ones. 
Extensive evaluations on two pioneering DVLMs, LLaDA-V and LaViDa, across multiple widely used benchmarks demonstrate that our proposed RedVTP can significantly improve inference efficiency while maintaining accuracy, and even enhance accuracy on certain benchmarks. Our work provides a new perspective for accelerating the inference of DVLMs.


For the future work, we will explore the orthogonality of RedVTP and any KV cache methods that are proposed in Diffusion Language Models (DLMs), although the effectiveness of these techniques have not been explored in DVLMs. Our RedVTP should be orthogonal to KV cache related methods. Intuitively, we can simply prune the corresponding KV cache of the less important visual tokens. Therefore, we claim that the two methods can be applied jointly without interference.

%% file: sec/X_suppl.tex
\clearpage
\setcounter{page}{1}
\maketitlesupplementary

\section{Appendix}
\label{sec:rationale}
\subsection{Algorithm}
\label{subsec:alg}
The algorithmic details of our RedVTP are presented in Algorithm~\ref{alg:vis_token_importance}.

\begin{table*}[ht]
\centering
\resizebox{0.7\textwidth}{!}{%
\setlength{\tabcolsep}{8pt}
\renewcommand{\arraystretch}{1}
\begin{tabular}{llccc}
\toprule
\multicolumn{5}{c}{\textbf{Efficiency}} \\
\textbf{Model \& Method} &  & RealworldQA~\cite{xai2024grok1.5v} & DocVQA~\cite{mathew2021docvqadatasetvqadocument} & InfoVQA~\cite{mathew2021infographicvqa} \\
\midrule
\multirow{2}{*}{Ours}
 & Latency (s/sample)~$\downarrow$       & \textbf{2.746} & \textbf{9.725} & \textbf{9.601} \\
 & Throughput (tok/s)~$\uparrow$         & \textbf{0.364}  & \textbf{3.187} & \textbf{3.229} \\
\cmidrule(l){2-5}
\multirow{2}{*}{PP} 
 & Latency (s/sample)~$\downarrow$       & 2.746 & 13.577 & 13.419 \\
 & Throughput (tok/s)~$\uparrow$         & 0.364  & 2.283 & 2.310 \\
\bottomrule
\end{tabular}%
}
\caption{Comparison of latency and throughput between our method and PP when applied to LLaDA-V}
\label{tab:efficiency_lavida}
\end{table*}

\begin{algorithm}[h]
\LinesNumbered
\caption{Response-driven Visual Token Pruning}
\label{alg:vis_token_importance}
\KwIn{
    Attention maps in DLM \( P_\theta(\cdot) \) at the \( 1 \)-st inference step $\mathbf{A}_1^{\mathrm{(l,h)}}$;
    Indices of masked response tokens after the \( 1 \)-st inference step $\mathcal{I}_1(M)$;
    Visual tokens $V \in \mathbb{R}^{N \times d}$; Indices of visual tokens $\mathcal{I}(V)$
}
\KwOut{
Retained visual tokens $V^{\text{keep}}\in \mathbb{R}^{N_r \cdot d}$
}
\SetAlgoLined

\textit{/* Masked token-guided importance score */}\\
$\bar{\mathbf{A}} = \frac{1}{H} \frac{1}{L} \sum_{h=1}^{H} \sum_{l=1}^{L} \mathbf{A}_1^{(l,h)}$\;
$\mathbf{S_1} = \frac{1}{|\mathcal{I}_1(M)|} \sum_{j \in \mathcal{I}_1(M)} \bar{\mathbf{A}}_{j, \mathcal{I}(V)}$ \;
\textit{/* Visual token pruning */}\\
$\mathcal{I}^{\text{keep}} = \texttt{Top}(\mathcal{I}(V), \mathbf{S}_1, r)$;\\
Retained visual tokens: $V^{\text{keep}} \gets V[\mathcal{I}^{\text{keep}}_{\text{sorted}}]$ \;
\Return{$V^{\text{keep}}$}
\end{algorithm}

\subsection{Efficiency Comparison Between Ours and Progressive Pruning}
\label{subsec:efficiencyPP}
As shown in Table~\ref{tab:efficiency_lavida}, our pruning method consistently achieves better inference efficiency compared to Progressive Pruning (PP). On average across these benchmarks, our method reduces latency by 18.94\% and improves throughput by 26.46\% relative to PP.